\documentclass[letterpaper,journal]{IEEEtran}

\usepackage{amsmath,amssymb,amsfonts}
\usepackage{graphicx}
\graphicspath{{figures/}}
\usepackage{booktabs}
\usepackage{textcomp}
\usepackage{stfloats}
\usepackage{url}
\usepackage{cite}
\usepackage{tikz}
\usetikzlibrary{positioning,arrows.meta}
\begin{document}

\title{Real-Time GPU-Accelerated Monte Carlo Evaluation of Safety-Critical AEB Systems Under Uncertainty}

\author{Akshay Karjol,~\IEEEmembership{Senior Member, IEEE,}
and Shadi Alawneh,~\IEEEmembership{Senior Member, IEEE}
\thanks{A. Karjol and S. Alawneh are with the Department of Electrical and Computer Engineering, Oakland University, Rochester, MI 48309 USA (e-mail: akshaykarjol@oakland.edu; shadialawneh@oakland.edu).}}

\markboth{}{}

\IEEEaftertitletext{%
  \vspace{-2.1\baselineskip}%
  \begingroup\setlength{\fboxsep}{2pt}\setlength{\fboxrule}{0.4pt}%
  \noindent
  \fbox{\begin{minipage}{\dimexpr\textwidth-2\fboxsep-2\fboxrule\relax}\footnotesize\noindent
  This work has been submitted to the IEEE for possible publication. Copyright may be transferred without notice, after which this version may no longer be accessible.%
  \end{minipage}}%
  \endgroup\vspace{0.55\baselineskip}%
}

\maketitle

\begin{abstract}
Automatic Emergency Braking (AEB) systems represent a safety-critical national interest, with the National Highway Traffic Safety Administration (NHTSA) Federal Motor Vehicle Safety Standard (FMVSS No. 127) requiring AEB in all new light vehicles sold in the United States by September 2029. However, production implementations frequently rely on deterministic stopping-distance or Time-to-Collision (TTC) thresholds that fail to capture uncertainty in sensing, road conditions, and vehicle dynamics. This paper presents a GPU-accelerated Monte Carlo framework for stochastic evaluation of emergency braking performance using a longitudinal vehicle model incorporating aerodynamic drag, road grade, brake actuator dynamics, and weight transfer effects. A one-thread-per-sample execution strategy exploits the independence of Monte Carlo rollouts, while deterministic CPU-generated sampling ensures bit-exact numerical consistency between CPU and GPU implementations. The framework is evaluated across four hardware platforms: two laptop GPUs (GTX 1650, RTX 5070) and two automotive-grade embedded platforms (Jetson Orin Nano, Jetson AGX Orin). Peak speedups of 54.57$\times$ are achieved. Real-time feasibility analysis with a complete AEB timing budget demonstrates that the Jetson AGX Orin can execute approximately 25,000 Monte Carlo samples within a 530 ms budget, enabling real-time probabilistic AEB evaluation as part of a complete embedded pipeline. Collision probability analysis demonstrates that a threshold set at the mean stopping distance of 79.2 m leaves a collision probability of approximately 47\%, while the conservative 1\% risk threshold requires a minimum safe headway of 111.1 m, an additional 31.9 m (40\%). These results provide quantitative guidance for risk-aware AEB threshold selection under the NHTSA final rule.
\end{abstract}


\section{Introduction}
\label{sec:introduction}

\IEEEPARstart{A}{utomatic} Emergency Braking (AEB) represents a safety-critical component of modern passenger vehicles and a significant national interest in automotive safety. AEB systems prevent or mitigate rear-end collisions by autonomously intervening when a driver fails to react in time. The National Highway Traffic Safety Administration (NHTSA) final rule (FMVSS No. 127), published May 2024, requires all new light vehicles sold in the United States to incorporate AEB systems by September 2029~\cite{NHTSA2024AEBFinalRule}. This final rule addresses a critical public safety concern. Rear-end collisions account for $29\%$ of all police-reported crashes in the United States~\cite{NHTSA2023TrafficSafetyFacts2021}. Meeting this requirement demands evaluation methodologies that are both comprehensive and computationally efficient. Manufacturers must validate system behavior across diverse environmental and operational uncertainties to ensure public safety and compliance with federal regulations.

While AEB is now standard in production vehicles, current evaluation approaches present significant limitations. Most systems rely on deterministic Time-to-Collision (TTC) thresholds or simplified stopping-distance formulas~\cite{van2012time,ISO22737} that assume ideal sensing and constant relative velocities. Berthelot \textit{et al.}~\cite{Berthelot2012IROSStochasticSituation} demonstrated that deterministic TTC assessment cannot handle perception uncertainties, leading to unrobust decisions or erroneous warnings. St\"{o}ckle \textit{et al.}~\cite{Stockle2018ITSC} observed that sensor measurement errors cause false interpretation of driving situations. Dahl \textit{et al.}~\cite{Dahl2023PUAwareThreatDetection} showed that prediction uncertainty affects threat detection in advanced driver assistance systems.

These limitations motivate probabilistic evaluation frameworks that propagate uncertainty through the decision process. Among available probabilistic methods, Monte Carlo simulation offers unique advantages over Polynomial Chaos Expansion, Gaussian Processes, and analytical bounds for AEB applications~\cite{Stockle2020SPM}. It handles arbitrary non-Gaussian distributions, nonlinear dynamics, and the discontinuities inherent in emergency braking without restrictive smoothness assumptions~\cite{Robert2004MonteCarloStatistical}. Section~II-D provides a detailed comparison. However, Monte Carlo methods are computationally demanding~\cite{Esler2012QMCGPU,Dai2025SafetyDominantSMPD}. Executing thousands of trajectory rollouts within AEB timing constraints is challenging on CPU architectures alone. GPU parallelization addresses this limitation by exploiting the statistical independence of Monte Carlo rollouts~\cite{owens2008gpu,aila2009understanding}.

While Monte Carlo simulation and GPU parallelization are individually established techniques, their combined evaluation for AEB stopping-distance uncertainty on automotive-grade embedded platforms has not been reported (Section~II-E). To address these challenges and support NHTSA-compliant AEB evaluation, this work makes the following contributions:
\begin{itemize}
\item \textbf{Braking model:} We develop a detailed longitudinal vehicle model that incorporates aerodynamic drag, road grade, weight transfer, and brake-actuator first-order dynamics.
\item \textbf{Deterministic CPU-GPU Monte Carlo framework with exact numerical consistency:} We implement a simulation architecture supporting both CPU and GPU execution with identical sampling and numerical procedures. This achieves bit-exact stopping-distance agreement across platforms.
\item \textbf{Comprehensive cross-platform performance evaluation on automotive-grade hardware:} We present a timing and scalability study across two laptop GPUs (GTX~1650, RTX~5070) and two automotive-grade embedded GPUs (Jetson Orin Nano, Jetson AGX Orin), achieving speedups ranging from $13.33\times$ to $54.57\times$.
\item \textbf{Collision probability analysis for risk-aware AEB intervention:} We derive collision probability curves $P(D_{\text{stop}}>H_0)$ from stopping-distance distributions, enabling quantitative risk-aware AEB threshold selection for multiple safety margins.
\end{itemize}

The results establish a scalable computational foundation for real-time uncertainty-aware AEB decision making and probabilistic braking risk estimation. The demonstrated cross-platform consistency and acceleration confirm viability across development (laptop) and deployment (embedded) environments.

The remainder of this paper is organized as follows. Section~II reviews related work in uncertainty quantification for AEB, GPU-accelerated Monte Carlo methods, and embedded computing for automotive applications. Section~III formulates the emergency braking problem and introduces the vehicle dynamics model. Section~IV describes the Monte Carlo framework and deterministic sampling approach. Section~V details the GPU implementation and parallelization strategy. Section~VI presents experimental results demonstrating numerical consistency, cross-platform performance, and real-time feasibility analysis. Sections~VII and~VIII conclude the paper and outline future work.

\section{Related Work}
\label{sec:related-work}

Uncertainty-aware AEB evaluation requires integrating three research domains: stochastic methodologies for automotive safety systems, GPU-accelerated simulation techniques, and automotive-grade embedded computing platforms. This section reviews prior work in each area (Sections~II-A through II-C), provides a systematic comparison of uncertainty quantification methods (Section~II-D), and positions the current contribution relative to existing literature (Section~II-E).

\subsection{Uncertainty Quantification in AEB Systems}
\label{sec:uq-aeb}

Traditional AEB systems rely on deterministic Time-to-Collision (TTC) thresholds~\cite{van2012time,ISO22737}. These assume ideal sensing conditions and constant relative velocities~\cite{Berthelot2012IROSStochasticSituation}. Real-world AEB performance depends on uncertainties in sensor measurements, road conditions, and vehicle dynamics~\cite{Stockle2018ITSC,Stockle2020SPM}. Recent work addresses these limitations through probabilistic approaches.

St\"{o}ckle \textit{et al.}~\cite{Stockle2018ITSC,Stockle2019IV} developed analytical methods for robust AEB design considering sensor measurement errors. They formulated the problem as an optimization over probabilistic quality measures. These analytical approaches provide closed-form solutions for Gaussian uncertainties but are limited to simplified scenarios. They cannot accommodate complex non-Gaussian uncertainty distributions. Building on similar analytical foundations, Leyrer \textit{et al.}~\cite{Leyrer2020IV} proposed the Orthogonal Worst-Case Distance (OWCD) approach as an efficient alternative to Monte Carlo simulation.

Norden \textit{et al.}~\cite{norden2023uncertainty} presented uncertainty-aware stopping-distance prediction using stochastic simulation. Their work focused on high-level scenario generation rather than computational infrastructure for real-time embedded evaluation. The review by St\"{o}ckle \textit{et al.}~\cite{Stockle2020SPM} surveys robust design methodologies for automated vehicular safety systems. It compares analytical, Monte Carlo, and worst-case approaches but does not address GPU acceleration or embedded deployment.

\subsection{GPU Acceleration for Monte Carlo Methods}
\label{sec:gpu-mc}

GPU-accelerated Monte Carlo methods have been studied in high-performance computing~\cite{owens2008gpu,aila2009understanding}. These methods demonstrate significant speedups for parallel workloads. Esler \textit{et al.}~\cite{Esler2012QMCGPU} achieved $10$--$15\times$ speedups for quantum Monte Carlo simulations on desktop GPUs. Navarro \textit{et al.}~\cite{Navarro2015MultiGPUExchangeMC} demonstrated multi-GPU Monte Carlo with $99\%$ efficiency for physics simulations. These works focus on desktop or datacenter GPU platforms and do not address automotive embedded system constraints.

In automotive applications, Dai \textit{et al.}~\cite{Dai2025SafetyDominantSMPD} developed safety-dominant stochastic model predictive decision-making using GPU-accelerated Monte Carlo for obstacle avoidance. Their approach samples uncertain obstacle trajectories within a closed-loop model predictive control framework to compute optimal evasive maneuvers, achieving real-time performance on an industrial workstation equipped with an NVIDIA RTX~A6000 GPU. However, their work does not address automotive-grade embedded platforms or AEB stopping-distance uncertainty quantification.

\subsection{Embedded Computing for Automotive Applications}
\label{sec:embedded-auto}

Embedded GPU platforms for automotive applications require real-time processing within strict power and thermal constraints~\cite{BorregoCarazo2020ResourceConstrainedMLADAS,Mazzocchetti2019AutoGPUPerf}. Borrego-Carazo \textit{et al.}~\cite{BorregoCarazo2020ResourceConstrainedMLADAS} provide a systematic review of resource-constrained machine learning for ADAS. This covers hardware from MPSoC CPUs to embedded GPUs including NVIDIA's Jetson family. Farooq \textit{et al.}~\cite{Farooq2023ThermalEmbeddedGPU} evaluated thermal imaging on Jetson Nano and Xavier NX for vehicular assistance systems. They achieved $11$~fps and $60$~fps through TensorRT-based optimization.

Mazzocchetti \textit{et al.}~\cite{Mazzocchetti2019AutoGPUPerf} analyzed performance optimization for the Jetson TX2 platform. Fickenscher \textit{et al.}~\cite{Fickenscher2018DATEOccupancyGrid} demonstrated GPU-accelerated occupancy grid mapping on the Jetson~K1, achieving $2.5$--$4.5\times$ speedups for ADAS applications. These works establish the Jetson family as a viable platform for automotive embedded computing. However, they focus on perception and mapping tasks rather than uncertainty quantification or Monte Carlo simulation for safety-critical evaluation.

\subsection{Comparison with Alternative Approaches}
\label{sec:uq-comparison}

The choice of uncertainty quantification method for AEB evaluation involves trade-offs between computational efficiency, modeling flexibility, and accuracy. This subsection compares Monte Carlo simulation against Polynomial Chaos Expansion (PCE), Gaussian Processes (GP), and analytical methods.

\textit{Polynomial Chaos Expansion (PCE):} PCE represents uncertain quantities through orthogonal polynomial bases and achieves spectral convergence for smooth systems~\cite{Xiu2002WienerAskeyPC}. Yang \textit{et al.}~\cite{Yang2020ACCGPCPropulsion} demonstrated over two orders of magnitude speedup relative to CPU-based Monte Carlo. However, PCE requires smooth dynamics and exhibits poor convergence for discontinuous systems. Emergency braking involves brake engagement, tire saturation, and actuator dynamics that violate smoothness assumptions. PCE also suffers from exponential complexity growth with increasing uncertainty dimensionality.

\textit{Gaussian Process (GP) Models:} GP-based approaches provide Bayesian uncertainty estimates for automotive prediction and control~\cite{Rasmussen2006GPML,Liu2022IVGPMPCOvertaking}. Winkelmann \textit{et al.}~\cite{Winkelmann2022ProbabilisticMetamodels} showed that GPs offer reliable uncertainty estimates but suffer from cubic computational complexity and restrictive Gaussian noise assumptions.

\textit{Analytical Methods:} Analytical approaches derive closed-form performance bounds using worst-case analysis and probabilistic constraints~\cite{Stellet2016IVBoundsAEB}. The OWCD framework~\cite{Stockle2018ITSC,Stockle2019IV,Stockle2020SPM} achieves computational efficiency but relies on restrictive assumptions. These include Gaussian sensor noise, simplified dynamics, and specific mathematical structure. These limitations reduce applicability to production AEB systems.

\textit{Why Monte Carlo?:} Monte Carlo simulation propagates uncertainty through nonlinear, discontinuous dynamics~\cite{Lefevre2014Survey,norden2023uncertainty}. Its convergence rate is independent of dimensionality. Its parallel structure enables near-linear speedup on GPUs~\cite{owens2008gpu,aila2009understanding,Esler2012QMCGPU,Navarro2015MultiGPUExchangeMC}. GPU acceleration eliminates Monte Carlo's traditional computational disadvantage while preserving modeling flexibility.

\subsection{Positioning of This Work}
\label{sec:positioning}

This work addresses the gap at the intersection of uncertainty-aware AEB evaluation, GPU-accelerated Monte Carlo simulation, and embedded automotive computing. Unlike prior analytical approaches~\cite{Stockle2018ITSC,Stockle2019IV,Stockle2020SPM}, the proposed framework handles arbitrary nonlinear dynamics and non-Gaussian uncertainties through Monte Carlo sampling. Unlike general GPU Monte Carlo studies~\cite{Esler2012QMCGPU,Navarro2015MultiGPUExchangeMC} focused on desktop or datacenter platforms, this work evaluates laptop and automotive-grade embedded GPUs.

The work by Dai \textit{et al.}~\cite{Dai2025SafetyDominantSMPD} represents the closest prior art in GPU-accelerated Monte Carlo for automotive applications; however, it addresses a fundamentally different problem. Their approach samples uncertain \textit{obstacle trajectories} within a closed-loop model predictive control framework to compute optimal evasive maneuvers. In contrast, this work propagates \textit{ego-vehicle parameter uncertainty} (friction, mass, road grade) through open-loop braking dynamics to characterize stopping-distance distributions for AEB threshold calibration. Furthermore, while their evaluation relies on a high-power workstation GPU (RTX~A6000) unsuitable for vehicle deployment, this work demonstrates feasibility on automotive-grade embedded platforms (Jetson Orin family).

To our knowledge, no prior work has demonstrated cross-platform Monte Carlo evaluation spanning laptop and embedded automotive GPUs for AEB stopping-distance uncertainty quantification with explicit timing budget validation and bit-exact numerical reproducibility.

\section{Problem Formulation}
\label{sec:problem-formulation}

This work evaluates the stopping-distance distribution of a passenger vehicle undergoing emergency braking under uncertainty. Braking performance depends on initial speed and variations in vehicle parameters and road conditions. Monte Carlo simulation propagates these uncertainties through a vehicle-dynamics model. This section formalizes the braking problem and introduces the physics-based vehicle dynamics model.

\subsection{State and Control Definition}
\label{sec:state-control}

We consider a point-mass longitudinal vehicle model augmented with a braking actuator state. The system state is defined as
\begin{equation}
\mathbf{x}(t) =
\begin{bmatrix}
x(t) \\
v(t) \\
a_{\text{brake,actual}}(t)
\end{bmatrix},
\end{equation}
where $x(t)$ denotes the longitudinal position, $v(t)$ denotes the vehicle speed, and $a_{\text{brake,actual}}(t)$ denotes the realized braking deceleration after actuator dynamics. The control input to the system is the commanded braking deceleration,
\begin{equation}
a_{\text{brake,cmd}}(t),
\end{equation}
which represents the deceleration request generated by the AEB controller upon activation.

The system is initialized with $x(0)=0$, $v(0)=v_0$ (sampled from a prescribed uncertainty distribution), and $a_{\text{brake,actual}}(0)=0$. The braking command is issued at $t=0$. Actuator dynamics cause the realized braking to build up with a $150~\text{ms}$ time constant, representative of production hydraulic braking systems~\cite{Wong2022GroundVehicles}. Braking continues until $v(t)=0$. The stopping distance is defined as
\begin{equation}
D_{\text{stop}} = x(t_{\text{stop}}),
\end{equation}
where $t_{\text{stop}}$ is the time at which the vehicle reaches zero speed.

\subsection{Longitudinal Vehicle Dynamics}
\label{sec:longitudinal-dynamics}

Longitudinal acceleration during braking is governed by aerodynamic drag, gravitational effects due to road grade, and the realized braking deceleration~\cite{Wong2022GroundVehicles,Rajamani2012VehicleDynamicsControl}. The longitudinal force balance yields
\begin{equation}
\frac{dv}{dt} = a_{\text{total}}(v,\theta)
= a_{\text{brake,actual}}
- \frac{c_a}{m} v^2
- g\sin(\theta),
\end{equation}
where $c_a = \frac{1}{2}\rho C_d A_f$ denotes the lumped aerodynamic drag coefficient, $\rho$ denotes air density, $C_d$ denotes the aerodynamic drag coefficient, $A_f$ denotes the frontal area, $m$ denotes the vehicle mass, $g$ denotes the gravitational constant, and $\theta$ denotes the road-grade angle. The quadratic term models aerodynamic resistance. The grade term represents uphill or downhill effects.

To represent the physical limitations on braking capability, including weight transfer to the front axle during deceleration~\cite{Wong2022GroundVehicles,Rajamani2012VehicleDynamicsControl}, the commanded braking deceleration is saturated at
\begin{equation}
a_{\max} = \min\left(\mu g\cos\theta,\; g\cos\theta\,\frac{l_f}{h}\right),
\end{equation}
where $\mu$ denotes the tire-road friction coefficient, $l_f$ denotes the distance from the center of gravity to the front axle, and $h$ denotes the center-of-gravity height. The first term is the tire-road friction limit, and the second is the weight-transfer limit at which the rear axle is fully unloaded. This formulation provides a numerically stable representation of load transfer effects.

\subsection{Brake Actuator Dynamics}
\label{sec:actuator-dynamics}

Production braking systems do not respond instantaneously. To capture the rise-time behavior of hydraulic or electromechanical brake actuators, a first-order lag model is included~\cite{Franklin2019FeedbackControl},
\begin{equation}
\frac{d}{dt}a_{\text{brake,actual}} =
\frac{1}{\tau}\left(a_{\text{brake,cmd}} - a_{\text{brake,actual}}\right),
\end{equation}
where $\tau$ denotes the actuator time constant. This state captures the transient mismatch between commanded and realized deceleration and contributes significantly to stopping distance.

\subsection{Complete System Dynamics}
\label{sec:complete-dynamics}

Combining the above relationships, the complete system evolves according to
\begin{align}
\frac{dx}{dt} &= v, \\
\frac{dv}{dt} &= a_{\text{total}}(v,\theta), \\
\frac{d}{dt}a_{\text{brake,actual}} &=
\frac{1}{\tau}\left(a_{\text{brake,cmd}} - a_{\text{brake,actual}}\right).
\end{align}
These nonlinear ordinary differential equations form the basis of the Monte Carlo simulation framework described in Section~IV, where the numerical integration scheme and its suitability for GPU implementation are discussed in detail.

\subsection{Model Scope and Limitations}
\label{sec:model-limitations}

The point-mass longitudinal model employed in this work provides an appropriate abstraction for braking-distance evaluation. However, several limitations should be acknowledged:

\begin{enumerate}
\item \textbf{AEB activation:} The analysis begins at AEB activation and assumes a stationary lead vehicle. Perception errors, cut-in maneuvers, occlusion, and multi-target interactions that determine when AEB activates are not modeled. This is appropriate for evaluating braking performance after activation but would require extension for end-to-end AEB evaluation.
\item \textbf{Lateral dynamics:} Lateral motion, yaw dynamics, and combined slip effects are not modeled. This is appropriate for straight-line emergency braking but would require extension for curved-road or evasive maneuvers.
\item \textbf{Tire slip dynamics:} The friction coefficient $\mu$ is treated as a lumped parameter rather than modeling the full tire slip curve. This simplification does not capture ABS cycling behavior explicitly.
\item \textbf{Multi-body effects:} Pitch dynamics and suspension compression during braking are not explicitly modeled; their effects are approximated through the weight-transfer formulation.
\item \textbf{Thermal effects:} Brake fade due to thermal degradation is not modeled, which may underestimate stopping distances under repeated heavy braking.
\end{enumerate}

These simplifications are consistent with prior work on uncertainty-aware AEB evaluation~\cite{norden2023uncertainty,Stockle2018ITSC,Stockle2019IV,Stockle2020SPM} and are appropriate for the computational performance focus of this study. Extension to higher-fidelity models (e.g., full vehicle dynamics or detailed tire models) is straightforward within the Monte Carlo framework and would primarily affect per-sample computation time rather than the parallelization strategy.

\textit{Objective Clarification:} This work demonstrates computational feasibility of uncertainty-aware braking evaluation under realistic nonlinear dynamics. It does not replace high-fidelity multi-body vehicle simulators. This fidelity level enables large-scale Monte Carlo studies on automotive-grade embedded platforms.

\section{Monte Carlo Framework}
\label{sec:mc-framework}

This section describes the Monte Carlo simulation framework used to propagate parameter uncertainty through the nonlinear braking dynamics. Each Monte Carlo rollout corresponds to one realization of model parameters and external conditions. It produces an independent stopping-distance outcome. Repeated execution across many samples yields an empirical distribution that characterizes braking performance under uncertainty. This section details the uncertainty representation, sampling procedure, and simulation workflow.

\subsection{Uncertainty Representation}
\label{sec:uncertainty-rep}

Several model parameters are treated as random variables: initial speed, road grade, aerodynamic drag coefficient, vehicle mass, and tire-road friction coefficient. Let $\boldsymbol{\xi}$ denote the vector of random parameters sampled from prescribed probability distributions. For each realization $\boldsymbol{\xi}^{(i)}$, the system dynamics are integrated from the initial state until the vehicle stops. This produces an independent stopping-distance sample
\begin{equation}
D_{\text{stop}}^{(i)} = x^{(i)}\!\left(t_{\text{stop}}^{(i)}\right).
\end{equation}
The collection $\{D_{\text{stop}}^{(i)}\}_{i=1}^{N}$ forms the stopping-distance distribution used to evaluate numerical consistency between CPU and GPU execution and to assess computational performance.

\subsection{Simulation Parameters and Uncertainty Distributions}
\label{sec:sim-params}

All simulations use a fixed integration time step $\Delta t = 0.001~\text{s}$, maximum simulation horizon $T_{\max}=10~\text{s}$, and commanded braking deceleration $a_{\text{brake,cmd}}=-6.0~\text{m/s}^2$. Nominal vehicle parameters are: mass $m=1500~\text{kg}$, lumped aerodynamic drag coefficient $c_a=0.3~\text{kg/m}$, tire-road friction coefficient $\mu=0.8$, center-of-gravity height $h=0.5~\text{m}$, center-of-gravity to front axle distance $l_f=1.35~\text{m}$, and brake-actuator time constant $\tau=0.15~\text{s}$. Road grade is nominally $\theta_{\text{road}}=0$. The $1~\text{ms}$ step ensures consistency across platforms. The framework supports coarser integration steps to improve embedded real-time feasibility.

To model uncertainty, several parameters are treated as independent normal random variables:
\begin{align}
v_0 &\sim \mathcal{N}(30.0,\;2.0^2)\ \text{m/s},\\
\mu &\sim \mathcal{N}(0.8,\;0.1^2),\\
\theta_{\text{road}} &\sim \mathcal{N}(0.0,\;0.05^2)\ \text{rad},\\
m &\sim \mathcal{N}(1500,\;100^2)\ \text{kg},\\
c_a &\sim \mathcal{N}(0.3,\;0.05^2)\ \text{kg/m}.
\end{align}

\textit{Distribution Choice Justification:} Normal distributions are selected for tractability and interpretability. Means represent typical operating conditions. Standard deviations capture realistic operational variability. The chosen standard deviations (e.g., $\pm 2~\text{m/s}$ for initial speed and $\pm 0.1$ for friction) represent moderate uncertainty levels consistent with sensor measurement errors and environmental variability in production systems~\cite{Stockle2018ITSC,Stockle2020SPM}.

Random samples are generated using a fixed seed to ensure deterministic correspondence between CPU and GPU execution.

\subsection{Sampling Procedure}
\label{sec:sampling-proc}

For each Monte Carlo rollout, an independent parameter vector $\boldsymbol{\xi}^{(i)}$ is drawn from the associated distributions. Random number generation is performed entirely on the CPU using a fixed seed, and all sampled parameter vectors are transferred to the GPU before simulation. This ensures that CPU and GPU executions operate on identical inputs, enabling deterministic one-to-one comparison of stopping-distance results.

\subsection{Simulation Workflow}
\label{sec:workflow}

Fig.~\ref{fig:methodology} illustrates the overall methodology, and Fig.~\ref{fig:algorithm} summarizes the per-sample workflow.

\begin{figure}[!t]
\centering
\begin{tikzpicture}[
  node distance=3mm,
  box/.style={draw, rounded corners=2pt, align=center, font=\footnotesize,
              text width=7.4cm, inner sep=2.5pt},
  half/.style={draw, rounded corners=2pt, align=center, font=\footnotesize,
               text width=3.45cm, inner sep=2.5pt},
  arr/.style={-{Stealth[length=2mm]}, thick}]
\node[box] (s1) {CPU: sampling of $\boldsymbol{\xi}^{(i)}$, $i=1,\dots,N$ (fixed seed)};
\node[half, below=4mm of s1.south east, anchor=north east] (g1) {Host-to-device copy of sampled parameters};
\node[half, below=of g1] (g2) {GPU: one thread per sample, RK4 rollout};
\node[half, below=of g2] (g3) {Device-to-host copy of $D_{\text{stop}}^{(i)}$};
\node[half, left=0.32cm of g2] (c1) {CPU: sequential RK4 rollouts on the same samples};
\node[box, anchor=north] (s4) at ([yshift=-4mm]s1 |- g3.south) {Per-sample CPU--GPU consistency check of $D_{\text{stop}}^{(i)}$};
\node[box, below=of s4] (s5) {Stopping-distance distribution and $P(D_{\text{stop}} > H_0)$};
\draw[arr] (s1.south -| g1) -- (g1);
\draw[arr] (g1) -- (g2);
\draw[arr] (g2) -- (g3);
\draw[arr] (g3) -- (g3 |- s4.north);
\draw[arr] (s1.south -| c1) -- (c1);
\draw[arr] (c1) -- (c1 |- s4.north);
\draw[arr] (s4) -- (s5);
\end{tikzpicture}
\caption{Overview of the GPU-accelerated Monte Carlo methodology.}
\label{fig:methodology}
\end{figure}

\begin{figure}[!t]
\centering
\fbox{\begin{minipage}{0.94\columnwidth}
\footnotesize
\textbf{Algorithm 1:} Monte Carlo stopping-distance evaluation\\[2pt]
\textbf{Input:} $\{\boldsymbol{\xi}^{(i)}\}_{i=1}^{N}$, $a_{\text{brake,cmd}}$, $\Delta t$, $T_{\max}$\\
\textbf{Output:} $\{D_{\text{stop}}^{(i)}\}_{i=1}^{N}$
\begin{enumerate}
\setlength{\itemsep}{0pt}
\item \textbf{for} each sample $i$ in parallel (one GPU thread per sample) \textbf{do}
\item \quad Load the sampled parameter vector $\boldsymbol{\xi}^{(i)}$ from global memory.
\item \quad Initialize $\mathbf{x}(0) = [0,\; v_0^{(i)},\; 0]^{\top}$ and $k \leftarrow 0$.
\item \quad \textbf{while} $v > 0$ \textbf{and} $k < T_{\max}/\Delta t$ \textbf{do}
\item \quad\quad Advance $\mathbf{x}$ by one RK4 step of size $\Delta t$, with $a_{\text{brake,cmd}}$ saturated at $a_{\max}$.
\item \quad\quad $v \leftarrow \max(0, v)$, $k \leftarrow k+1$.
\item \quad Write $D_{\text{stop}}^{(i)} = x$ to the output array.
\end{enumerate}
\end{minipage}}
\caption{Pseudocode of the per-sample Monte Carlo workflow. The CPU baseline executes the same steps sequentially.}
\label{fig:algorithm}
\end{figure}

Given a sampled parameter vector $\boldsymbol{\xi}^{(i)}$, the simulator integrates the nonlinear dynamics until the vehicle reaches zero speed. A fixed-step fourth-order Runge-Kutta (RK4) integrator advances the system state with step size $\Delta t$~\cite{Butcher2016NumericalODE}. RK4 was selected over lower-order methods (e.g., forward Euler) and higher-order adaptive schemes (e.g., Dormand-Prince) as RK4 achieves $\mathcal{O}(\Delta t^5)$ local truncation error with four function evaluations per step~\cite{Butcher2016NumericalODE}, whereas Euler's $\mathcal{O}(\Delta t^2)$ error would require impractically small steps~\cite{Butcher2016NumericalODE}. Adaptive methods introduce variable workloads that complicate GPU parallelization~\cite{owens2008gpu,Esler2012QMCGPU}.

The integration sequence proceeds deterministically across all samples, with no data-dependent branching. This uniform control flow simplifies parallel execution on the GPU and supports reproducible comparisons between platforms~\cite{owens2008gpu,aila2009understanding}. To ensure consistency across platforms, the CPU and GPU implementations share identical numerical procedures, differing only in the manner of parallelization: the CPU performs rollouts sequentially, whereas the GPU assigns each rollout to an independent thread.

\subsection{CPU Baseline Implementation}
\label{sec:cpu-baseline}

The CPU implementation serves two purposes: (i) it provides a deterministic reference for validating GPU numerical accuracy, and (ii) it establishes a performance baseline. While straightforward and useful for verification, the sequential CPU implementation highlights the limited scalability of CPU-only Monte Carlo evaluation when large numbers of rollouts are required.

\subsection{Deterministic Comparison Between CPU and GPU}
\label{sec:deterministic-cmp}

To ensure a fair numerical comparison, both simulators operate on the same set of sampled model parameters. This design eliminates discrepancies arising from differences in sampling order or random-number generation. As demonstrated in Section~VI, this approach yields exact agreement between CPU and GPU stopping-distance outputs while revealing the substantial computational gains enabled by GPU parallelism.

\subsection{Computational Complexity}
\label{sec:complexity}

Both CPU and GPU implementations exhibit $\mathcal{O}(N)$ time complexity, where $N$ is the number of Monte Carlo samples. For the CPU, sequential execution yields total time
\begin{equation}
T_{\text{CPU}} = N\, t_{\text{sample}},
\end{equation}
where $t_{\text{sample}}$ is the per-rollout computation time. For the GPU, parallel execution across $P$ threads yields
\begin{equation}
T_{\text{GPU}} \approx \left\lceil \frac{N}{P}\right\rceil t_{\text{sample}} + t_{\text{overhead}},
\end{equation}
where $t_{\text{overhead}}$ accounts for kernel launch and memory transfer latency.

The speedup factor $S = T_{\text{CPU}}/T_{\text{GPU}}$ therefore depends on:
\begin{enumerate}
\item \textbf{Parallelism:} Larger $P$ increases speedup.
\item \textbf{Overhead amortization:} Larger $N$ amortizes fixed kernel launch costs.
\item \textbf{Memory bandwidth:} Parameter transfer time becomes significant for large $N$.
\item \textbf{Occupancy:} Utilization depends on thread-block configuration and resource usage.
\end{enumerate}
In practice, speedup increases with $N$ until GPU resources (compute units and memory bandwidth) saturate, after which speedup stabilizes. This behavior is confirmed experimentally in Section~VI-D.

\section{GPU Implementation}
\label{sec:gpu-implementation}

The Monte Carlo framework is suited for parallel execution because each rollout is independent~\cite{owens2008gpu,aila2009understanding}. This structural independence enables efficient mapping to single-instruction-multiple-thread (SIMT) architectures such as NVIDIA GPUs. This section describes the GPU execution model, memory organization, kernel structure, and measures to ensure numerical consistency with the CPU implementation.

\subsection{Parallelization Strategy}
\label{sec:parallelization}

Each Monte Carlo rollout is assigned to a single GPU thread. This one-thread-per-sample strategy eliminates synchronization across threads and preserves the deterministic control flow of the RK4 integration loop. Let $N$ denote the total number of Monte Carlo samples. The GPU kernel is launched with a one-dimensional grid of size
\begin{equation}
\text{gridSize} = \left\lceil \frac{N}{\text{blockSize}} \right\rceil,
\end{equation}
where a block size of $256$ threads is used throughout this work.

This configuration balances occupancy and register pressure across GPU architectures. Smaller block sizes reduce achievable occupancy. Larger block sizes increase register pressure and limit active warps per streaming multiprocessor~\cite{owens2008gpu}. The choice of $256$ threads per block is recommended for compute-bound kernels. It delivers robust performance across the GTX~1650 (Turing), RTX~5070 (Blackwell), and Jetson Orin (Ampere) platforms evaluated in this study.

\subsection{Memory Organization}
\label{sec:memory-org}

All random parameters are generated on the CPU and transferred to the GPU as contiguous arrays. Each parameter array resides in global memory. The initial velocity array and result array store scalar values at consecutive addresses, enabling coalesced memory access for these arrays~\cite{owens2008gpu,Esler2012QMCGPU}. The vehicle parameters use an array-of-structures layout where each thread loads its complete parameter set once at kernel launch; subsequent parameter accesses occur from registers.

Thread-local registers also store the state variables for each rollout: position, velocity, and actuator state. Each thread operates independently. No shared memory, atomic operations, or inter-thread communication mechanisms are required. This design eliminates race conditions and simplifies the kernel implementation.

\subsection{Kernel Structure}
\label{sec:kernel-structure}

Each GPU thread executes the same deterministic RK4 integration loop used in the CPU implementation. The kernel proceeds through the following sequence:
\begin{enumerate}
\item Load the sampled parameter vector $\boldsymbol{\xi}^{(i)}$ from global memory.
\item Initialize the state vector $\mathbf{x}(0)$ using the sampled initial speed.
\item Integrate the nonlinear dynamics using the RK4 scheme until the vehicle speed reaches zero.
\item Write the resulting stopping distance to an output array indexed by the thread identifier.
\end{enumerate}

Each thread solves a small, fixed-size system of ordinary differential equations, resulting in a uniform computational workload across threads within a warp. This uniformity minimizes branch divergence and contributes to stable and predictable performance across all evaluated GPU architectures~\cite{owens2008gpu}.

\subsection{Ensuring Numerical Consistency}
\label{sec:numerical-consistency}

A central requirement of this work is exact numerical correspondence between CPU and GPU outputs for every Monte Carlo sample. To satisfy this requirement, the GPU implementation uses:
\begin{itemize}
\item the same parameter samples generated deterministically on the CPU,
\item the same fixed-step RK4 numerical integrator,
\item identical floating-point precision,
\item identical termination conditions (specifically, $v \leq 0$ or $t \geq T_{\max}$).
\end{itemize}

The implementation avoids architecture-specific approximations, fused operations, and non-deterministic instructions. As demonstrated in Section~VI, the resulting stopping-distance values match the CPU outputs exactly for all hardware platforms evaluated.

\subsection{Computational Efficiency}
\label{sec:gpu-efficiency}

Although the RK4 solver within each thread proceeds sequentially, parallel execution across thousands of GPU threads yields substantial acceleration relative to the strictly sequential CPU baseline. The one-thread-per-sample mapping produces a nearly uniform workload across threads, with minimal warp divergence arising only from boundary checks and early termination when individual rollouts reach zero velocity.

Contiguous storage of initial velocity and result arrays enables coalesced global memory access, while exclusive use of thread-local registers eliminates the need for shared memory and synchronization. This design maintains high efficiency across heterogeneous GPU architectures. Laptop GPUs (GTX~1650, RTX~5070) and embedded GPUs (Jetson Orin family) all benefit from massive parallelism, with higher-end platforms achieving greater speedups due to increased compute units and memory bandwidth. Importantly, the implementation uses a fixed integration step of $1~\text{ms}$ for consistency, although coarser integration steps are supported and may further improve embedded real-time feasibility in future work.

\section{Results}
\label{sec:results}

The GPU-accelerated Monte Carlo framework was evaluated along four dimensions: (1) numerical consistency between CPU and GPU implementations, (2) computational efficiency across four hardware platforms, (3) collision probability analysis for risk-aware AEB decision-making, and (4) real-time feasibility for AEB applications. All GPU experiments used identical pre-generated parameter samples from the CPU implementation.

\subsection{Numerical Consistency}
\label{sec:numerical-consistency-results}

The GPU implementation must reproduce CPU stopping-distance outputs exactly. Numerical consistency was validated across all four hardware platforms (GTX~1650, RTX~5070, Jetson AGX Orin, Jetson Orin Nano) for all sample sizes tested. The maximum absolute deviation between CPU and GPU stopping distances is
\begin{equation}
\Delta_{\max} = 0.000000~\text{m},
\end{equation}
for all platforms and all values of $N$.

\begin{figure}[!t]
\centering
\includegraphics[width=\columnwidth]{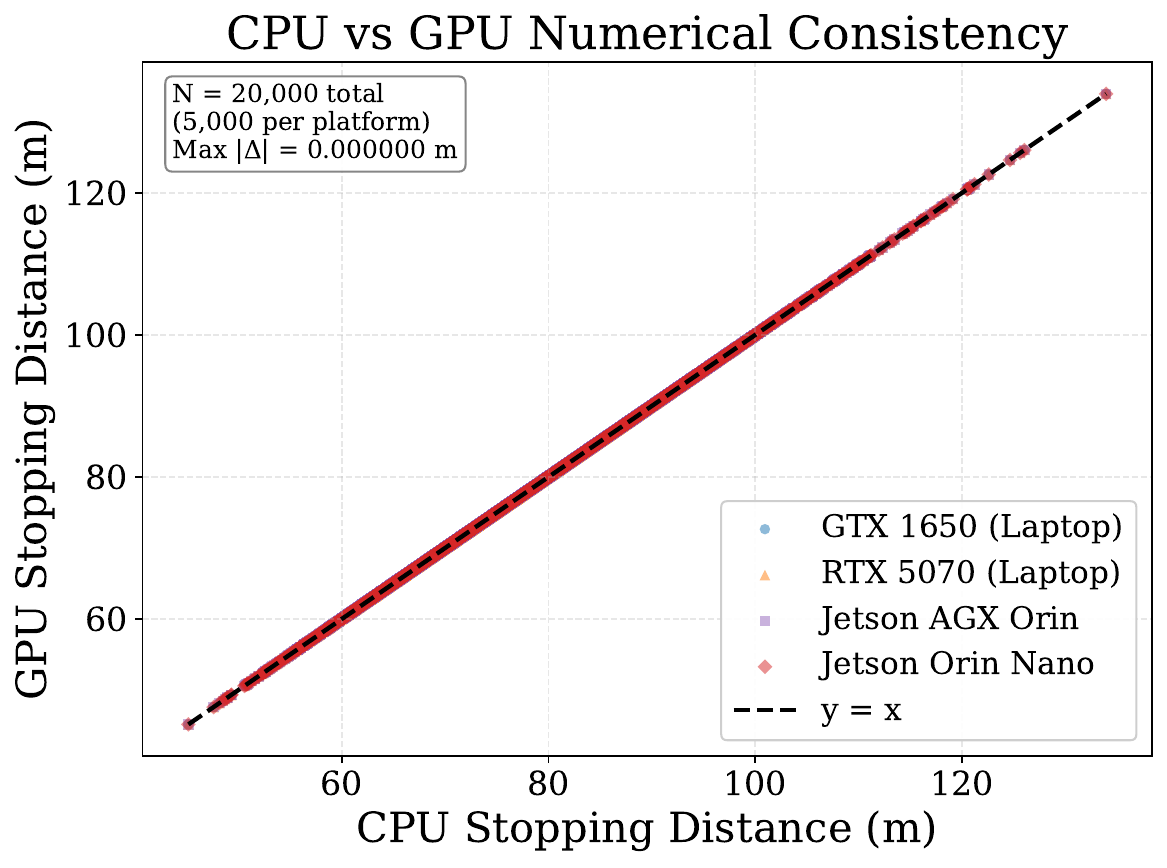}
\caption{CPU vs.\ GPU stopping-distance comparison across all four hardware platforms ($N=5{,}000$ per platform, $N=20{,}000$ total). Distinct markers identify each platform: GTX~1650 ($\circ$), RTX~5070 ($\triangle$), Jetson AGX Orin ($\square$), Jetson Orin Nano ($\diamond$). All points lie on the $y=x$ diagonal, confirming exact numerical agreement with maximum absolute deviation of $0.0~\text{m}$ for all platforms.}
\label{fig:cpu_gpu_consistency}
\end{figure}

Fig.~\ref{fig:cpu_gpu_consistency} illustrates this agreement for $N=5{,}000$ validation samples per platform across all four hardware configurations. All points from all platforms lie exactly on the $y=x$ diagonal, confirming that the GPU integrator preserves numerical fidelity regardless of GPU architecture. The four platforms are distinguished by marker shape: GTX~1650 (circle), RTX~5070 (triangle), Jetson AGX Orin (square), and Jetson Orin Nano (diamond).

\subsection{Stopping-Distance Distribution Under Uncertainty}
\label{sec:stopping-distribution}

\textit{Convergence Analysis:} Empirically, the stopping-distance distribution stabilizes for $N \geq 12{,}000$ samples under the selected uncertainty model. Table~\ref{tab:convergence} presents the convergence behavior of the Monte Carlo estimator as sample count increases. For $N \in [12\text{k}, 350\text{k}]$, the mean varies within $0.06~\text{m}$ ($79.19$--$79.25~\text{m}$) and standard deviation within $0.12~\text{m}$ ($12.09$--$12.21~\text{m}$), demonstrating that the Monte Carlo estimator has converged for practical purposes.

\begin{table}[!t]
\centering
\caption{Monte Carlo Convergence Analysis: Mean and Standard Deviation of Stopping Distance as a Function of Sample Count}
\label{tab:convergence}
\begin{tabular}{rcccc}
\toprule
$N$ & Mean $D_{\text{stop}}$ & Std.\ Dev.\ & $\Delta$ Mean & $\Delta$ Std.\ Dev.\ \\
    & (m) & (m) & (m) & (m) \\
\midrule
1{,}000   & 79.23 & 12.35 & +0.01 & +0.14 \\
4{,}000   & 79.20 & 12.11 & $-0.02$ & $-0.10$ \\
8{,}000   & 79.22 & 12.15 & 0.00 & $-0.06$ \\
\textbf{12{,}000} & \textbf{79.22} & \textbf{12.21} & \textbf{(baseline)} & \textbf{(baseline)} \\
25{,}000  & 79.20 & 12.16 & $-0.02$ & $-0.05$ \\
37{,}000  & 79.25 & 12.18 & +0.03 & $-0.03$ \\
65{,}000  & 79.24 & 12.15 & +0.02 & $-0.06$ \\
100{,}000 & 79.22 & 12.09 & 0.00 & $-0.12$ \\
150{,}000 & 79.22 & 12.13 & 0.00 & $-0.08$ \\
350{,}000 & 79.19 & 12.14 & $-0.03$ & $-0.07$ \\
\bottomrule
\end{tabular}
\end{table}

\begin{figure}[!t]
\centering
\includegraphics[width=\columnwidth]{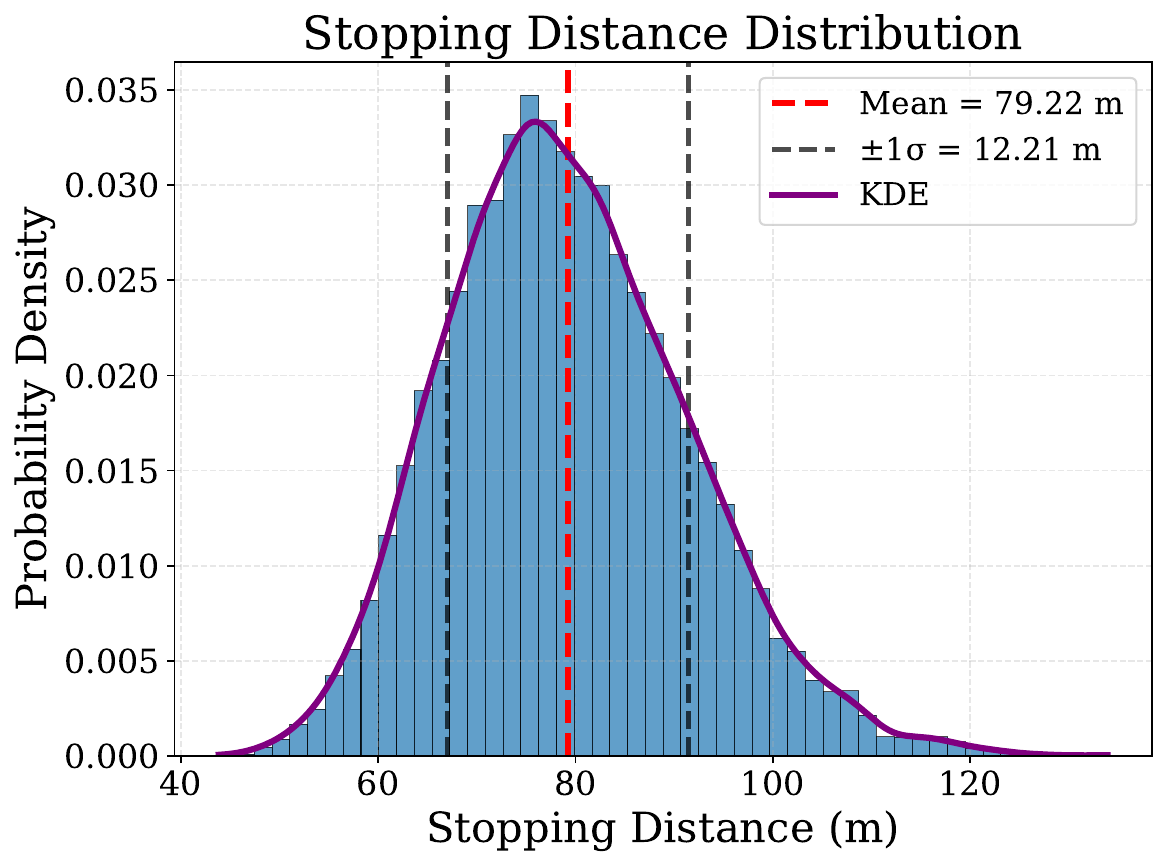}
\caption{Stopping-distance distribution from GPU Monte Carlo simulation at the convergence baseline. The distribution exhibits a mean of $79.22~\text{m}$ and standard deviation of $12.21~\text{m}$, capturing uncertainty in road grade, aerodynamic drag, friction, and initial speed.}
\label{fig:stopping_dist}
\end{figure}

Fig.~\ref{fig:stopping_dist} reports the stopping-distance distribution using the convergence baseline of $N=12{,}000$ Monte Carlo rollouts. The distribution is unimodal and slightly right-skewed with mean $79.22~\text{m}$ and standard deviation $12.21~\text{m}$. The observed range ($44$--$134~\text{m}$) reflects uncertainty in road grade, aerodynamic drag, friction, and initial speed.

\subsection{Collision Probability Analysis for Risk-Aware AEB}
\label{sec:collision-probability}

Practical AEB systems require actionable decision criteria. For a given initial headway $H_0$, the collision probability is defined as
\begin{equation}
P(\text{collision}\mid H_0) = P(D_{\text{stop}} > H_0),
\end{equation}
representing the fraction of uncertain scenarios in which the ego vehicle cannot stop before reaching the stationary lead vehicle.

Evaluating $P(D_{\text{stop}} > H_0)$ over a range of headway distances yields a collision-risk curve that informs AEB threshold selection. For the simulated braking scenario ($v_0 \sim \mathcal{N}(30, 2^2)~\text{m/s}$, $a_{\text{cmd}} = -6.0~\text{m/s}^2$), the minimum safe headways are:
\begin{itemize}
\item $5\%$ risk (aggressive): $H_0 = 100.6~\text{m}$
\item $1\%$ risk (conservative): $H_0 = 111.1~\text{m}$
\item $0.1\%$ risk (very conservative): $H_0 = 122.7~\text{m}$
\end{itemize}

\begin{figure}[!t]
\centering
\includegraphics[width=\columnwidth]{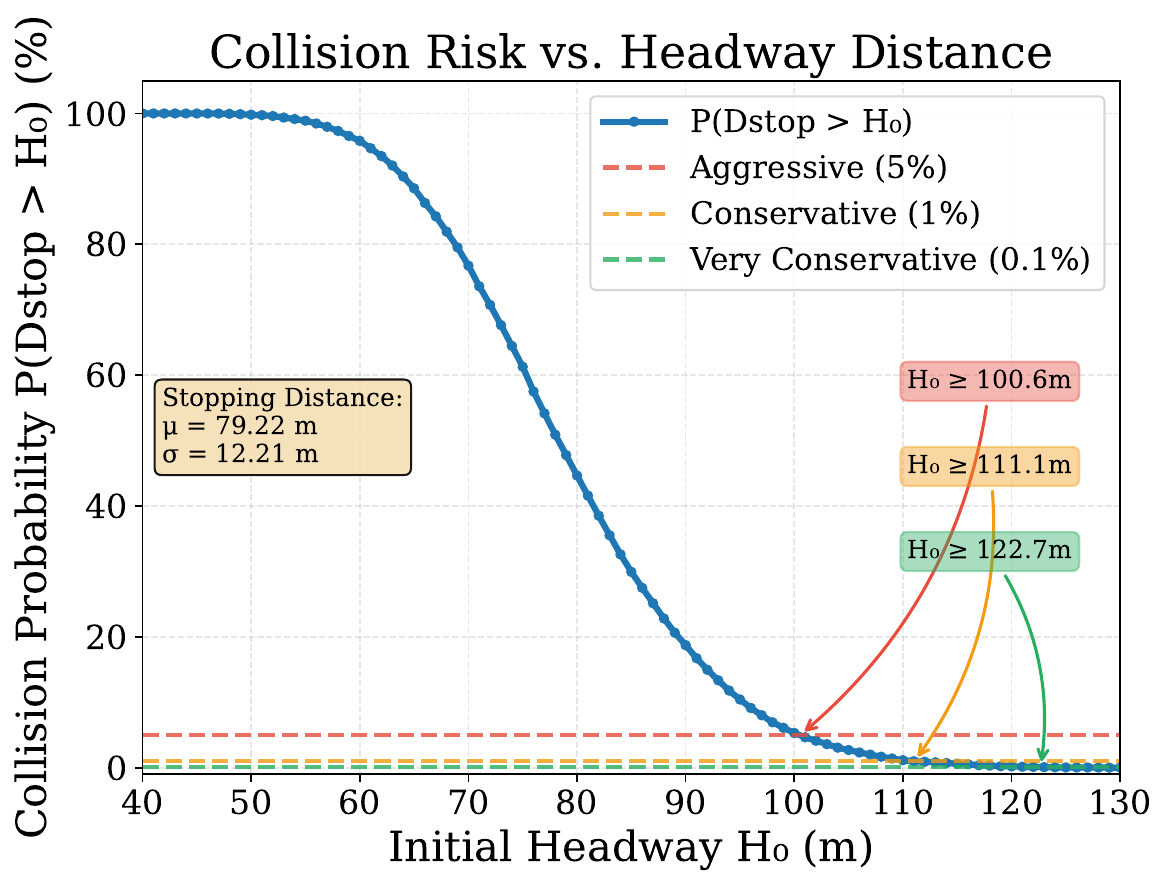}
\caption{Collision probability $P(D_{\text{stop}}>H_0)$ as a function of initial headway, computed from $N=12{,}000$ Monte Carlo samples. Horizontal dashed lines indicate risk tolerance thresholds: aggressive ($5\%$), conservative ($1\%$), and very conservative ($0.1\%$). Annotations show minimum safe headway for each threshold.}
\label{fig:collision_prob}
\end{figure}

These headway thresholds translate to TTC criteria via $\text{TTC}=H_0/v_{\text{rel}}$. At the mean initial speed of $30~\text{m/s}$ ($108~\text{km/h}$), the conservative $1\%$ risk threshold corresponds to $\text{TTC} \approx 3.7~\text{s}$. A threshold set at the mean stopping distance of $79.2~\text{m}$ leaves a collision probability of approximately $47\%$, while the conservative $1\%$ risk threshold requires an additional $31.9~\text{m}$ ($40\%$) of headway. Fig.~\ref{fig:collision_prob} shows the collision probability curve with annotated risk thresholds.

\subsection{Cross-Platform Performance Comparison}
\label{sec:cross-platform}

\begin{figure}[!t]
\centering
\includegraphics[width=\columnwidth]{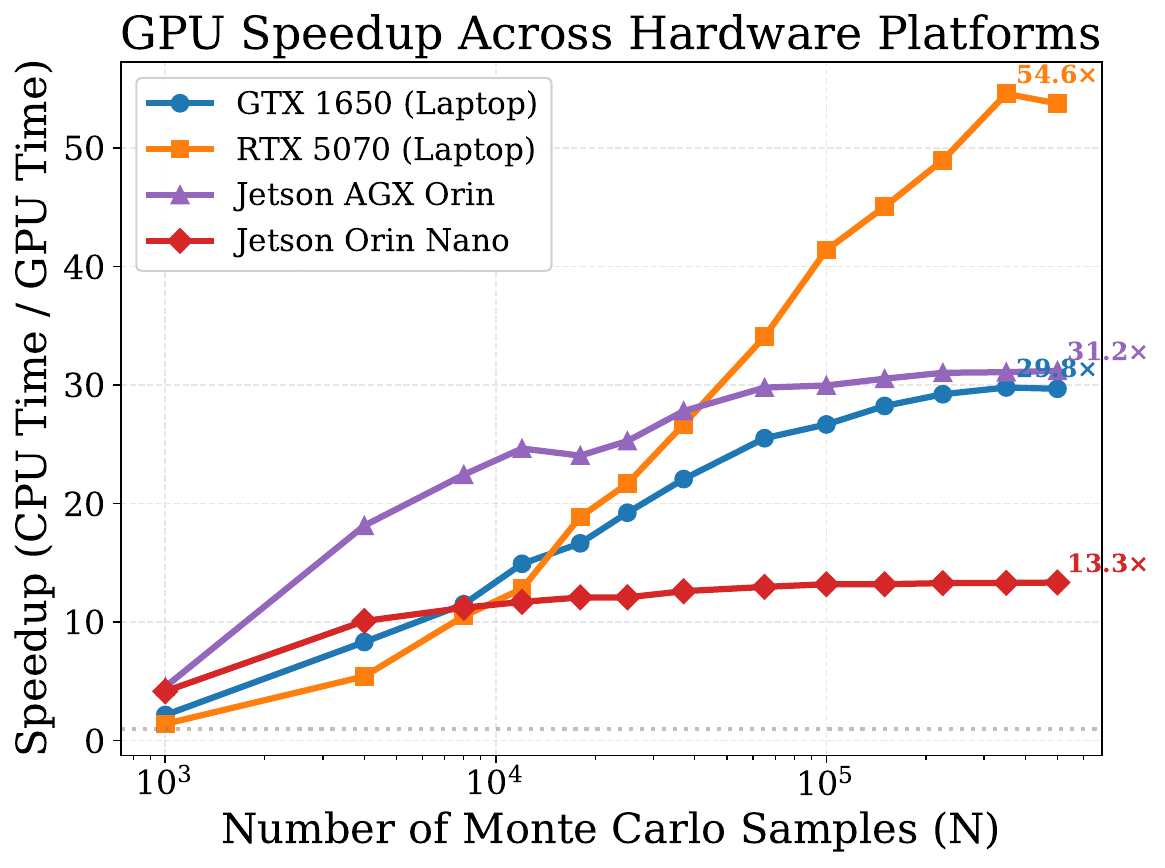}
\caption{GPU speedup relative to CPU baseline across four hardware platforms. Peak speedups range from $13.33\times$ (Jetson Orin Nano) to $54.57\times$ (RTX~5070). All platforms exhibit characteristic scaling behavior: kernel-overhead dominance at low $N$, linear scaling at moderate $N$, and saturation at high $N$.}
\label{fig:speedup}
\end{figure}

The framework was evaluated on four hardware platforms spanning development and deployment environments. Peak speedups relative to the CPU baseline range from $13.33\times$ (Jetson Orin Nano) to $54.57\times$ (RTX~5070). The Jetson AGX Orin achieves $31.20\times$ speedup on automotive-grade embedded hardware. Fig.~\ref{fig:speedup} presents the speedup scaling behavior across all platforms.

\subsection{Scaling Behavior with Sample Count}
\label{sec:scaling}

Performance scaling exhibits three characteristic regimes across all platforms:
\begin{enumerate}
\item \textbf{Kernel-overhead regime} ($N \leq 10{,}000$): speedup limited by launch and transfer overheads.
\item \textbf{Linear scaling regime} ($10{,}000 < N < 100{,}000$): speedup increases with parallel utilization.
\item \textbf{Saturation regime} ($N > 100{,}000$): speedup stabilizes as compute and memory bandwidth saturate.
\end{enumerate}

\begin{figure}[!t]
\centering
\includegraphics[width=\columnwidth]{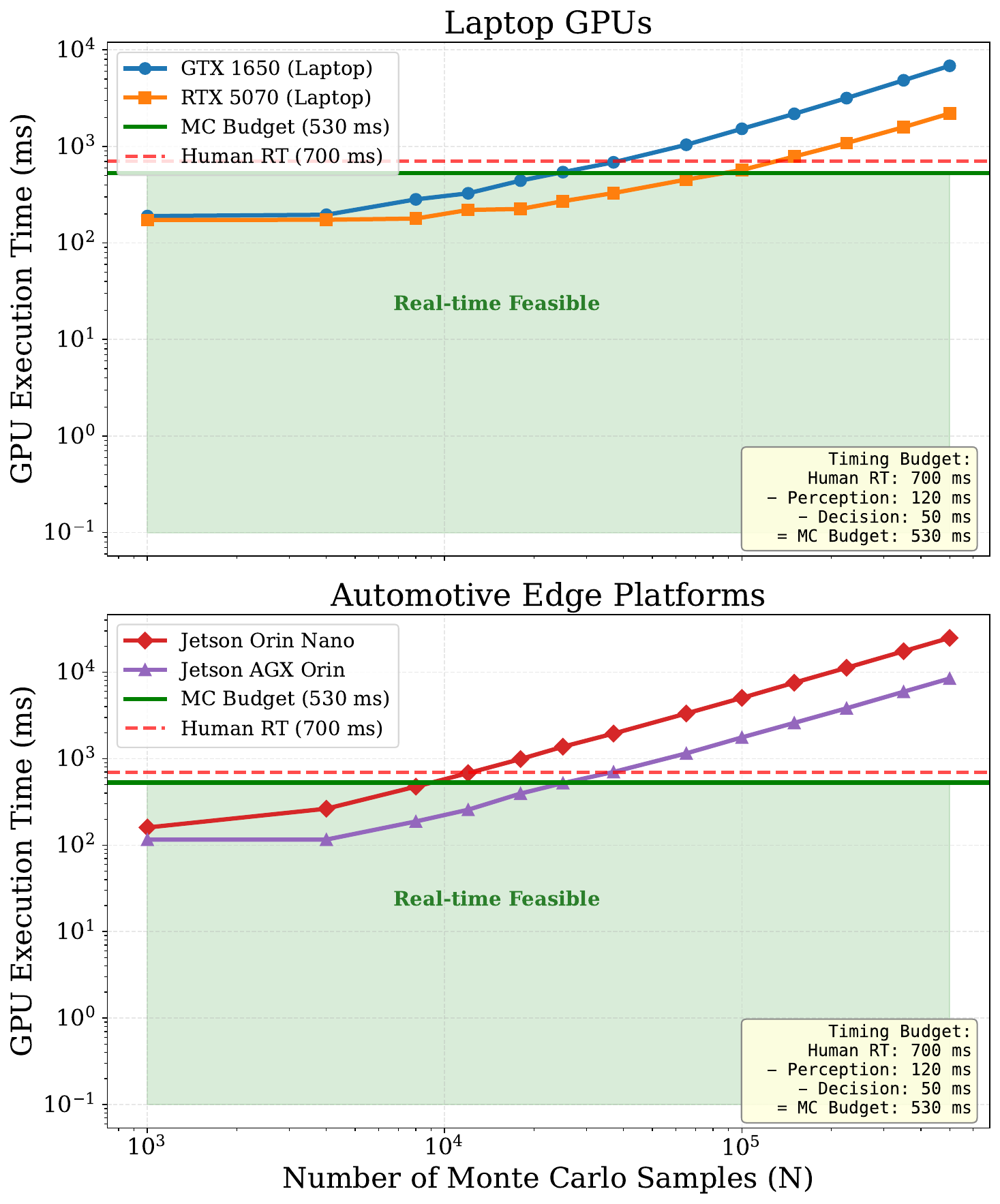}
\caption{GPU execution time vs. sample count across four platforms with AEB pipeline timing budget. The solid green line at $530~\text{ms}$ represents the Monte Carlo budget after accounting for perception ($120~\text{ms}$) and decision ($50~\text{ms}$) overhead. The dashed red line shows the total human reaction time ($700~\text{ms}$)~\cite{Green2000HowLongToStop}. The shaded region indicates feasibility for real-time AEB pipeline integration.}
\label{fig:exec_time}
\end{figure}

Fig.~\ref{fig:exec_time} illustrates the GPU execution time scaling across all platforms with reference timing thresholds for real-time AEB integration (also discussed in Section~\ref{sec:realtime}).

\subsection{Real-Time Feasibility Analysis}
\label{sec:realtime}

Human reaction time provides a practical upper bound for AEB system response, with total perception-to-actuation delays of approximately $700$--$750~\text{ms}$~\cite{Green2000HowLongToStop}. However, the complete AEB pipeline includes perception processing and decision logic in addition to Monte Carlo evaluation. Based on reported latencies in the literature, we adopt the following timing budget:

\begin{itemize}
\item \textbf{Perception processing:} $120~\text{ms}$ (upper bound for object detection, tracking, and sensor fusion~\cite{Paden2016SurveyMotionPlanning})
\item \textbf{Decision logic:} $50~\text{ms}$ (upper bound for risk evaluation on embedded platforms~\cite{Nilsson2017LaneChangeManeuvers})
\item \textbf{Monte Carlo budget:} $530~\text{ms}$ (remaining time for probabilistic evaluation)
\end{itemize}

This timing allocation represents one defensible reference configuration based on literature values. The framework's demonstrated scalability (Figs.~\ref{fig:speedup}--\ref{fig:exec_time}) enables adaptation to alternative timing budgets ranging from $400$--$800~\text{ms}$ depending on system architecture and component latencies.

\begin{figure}[!t]
\centering
\includegraphics[width=\columnwidth]{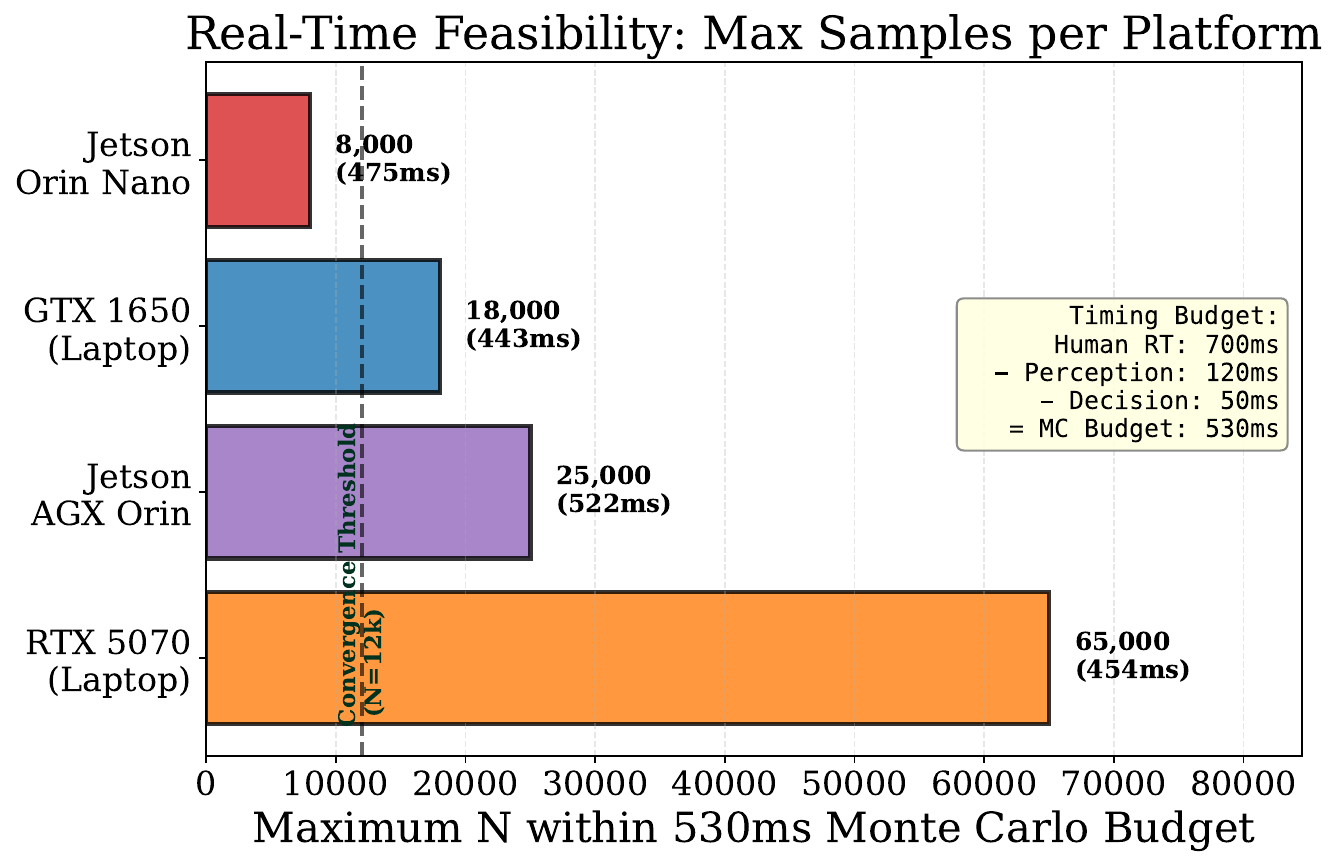}
\caption{Maximum Monte Carlo sample count achievable within the $530~\text{ms}$ Monte Carlo budget (from $700~\text{ms}$ total minus $120~\text{ms}$ perception and $50~\text{ms}$ decision overhead). The AGX Orin exceeds the $N=12{,}000$ convergence threshold (Table~\ref{tab:convergence}), while the Orin Nano ($N \approx 8{,}000$) remains within acceptable tolerance of converged values.}
\label{fig:realtime}
\end{figure}

With this Monte Carlo timing budget of 530 ms, the Jetson AGX Orin can execute up to $N \approx 25{,}000$ samples within the $530~\text{ms}$ Monte Carlo constraint (total pipeline: $\sim692~\text{ms}$), while the Jetson Orin Nano supports approximately $N \approx 8{,}000$ samples (total pipeline: $\sim645~\text{ms}$). The AGX Orin exceeds the $N = 12{,}000$ convergence threshold (Table~\ref{tab:convergence}). The Orin Nano falls below this threshold but remains within the tolerance range where mean and standard deviation deviate by at most $0.06~\text{m}$ from converged values, still providing statistically meaningful uncertainty characterization. Fig.~\ref{fig:realtime} summarizes the maximum achievable sample count for each platform within the Monte Carlo timing budget.

\subsection{Implications for AEB System Development}
\label{sec:implications}

The demonstrated numerical consistency and cross-platform acceleration enable a unified validation workflow: AEB algorithms can be developed and calibrated on laptop GPUs and deployed to embedded automotive platforms with bit-exact reproducibility. This capability simplifies regression testing, certification documentation, and compliance verification under the NHTSA AEB final rule.

\section{Conclusion}
\label{sec:conclusion}

This work presented a GPU-accelerated Monte Carlo framework for stochastic evaluation of emergency braking performance in AEB scenarios. The framework demonstrates practical feasibility on automotive-grade embedded platforms. It builds upon a longitudinal dynamics model incorporating aerodynamic drag, road-grade effects, weight transfer, and brake actuator dynamics. A one-thread-per-sample GPU execution strategy exploits the statistical independence of Monte Carlo rollouts. Deterministic sampling ensures CPU and GPU implementations produce numerically identical stopping-distance results.

Across all four hardware platforms evaluated (laptop GPUs: GTX~1650, RTX~5070; embedded GPUs: Jetson Orin Nano, Jetson AGX Orin), the proposed method reproduces CPU stopping-distance results with zero numerical deviation. Speedups range from $13.33\times$ (Jetson Orin Nano) to $29.79\times$ (GTX~1650), $31.20\times$ (Jetson AGX Orin), and $54.57\times$ (RTX~5070). Scaling analyses reveal expected performance regimes: kernel-overhead dominance at small sample counts, near-linear GPU throughput at moderate sample sizes, and saturation at large sample counts.

Real-time feasibility analysis with a complete AEB pipeline timing budget demonstrates that the Jetson AGX Orin can execute up to $N \approx 25{,}000$ Monte Carlo samples within the $530~\text{ms}$ Monte Carlo budget. This accounts for $120~\text{ms}$ perception and $50~\text{ms}$ decision overhead from the $700~\text{ms}$ human reaction-time threshold. Meaningful probabilistic AEB evaluation is achievable on automotive-grade embedded platforms as part of a complete processing pipeline.

Collision probability analysis demonstrates practical utility for risk-aware AEB decision-making. By computing $P(D_{\text{stop}}>H_0)$ across initial headway distances, the framework provides quantitative guidance for minimum safe headway selection. For the representative braking scenario, minimum safe headways range from $100.6~\text{m}$ (aggressive $5\%$ risk) to $122.7~\text{m}$ (very conservative $0.1\%$ risk). These translate to TTC thresholds for AEB system design. Collision probability curves are numerically identical across all evaluated platforms due to bit-exact stopping-distance agreement.

The demonstrated acceleration, cross-platform consistency, and risk-aware decision support establish a computational foundation for NHTSA-compliant AEB evaluation. The framework separates the dynamics model from the simulation backend and enforces deterministic CPU-GPU consistency. This enables uncertainty-aware AEB threshold calibration and probabilistic braking risk estimation under parameter uncertainty.


\section{Future Work}
\label{sec:future-work}

Several extensions would strengthen applicability to uncertainty-aware AEB design. This work demonstrates collision probability analysis $P(D_{\text{stop}}>H_0)$ for fixed headway scenarios. A next step is incorporation of dynamic lead-vehicle motion and time-varying headway within the Monte Carlo simulation loop. This extension would enable closed-loop evaluation where uncertainty in lead-vehicle behavior propagates through the braking trajectory.

Incorporating perception effects such as sensing latency, measurement noise, dropout, and occlusion would enhance realism. This would enable evaluation of AEB behavior under stochastic perception uncertainty. Kernel-level optimizations (shared-memory buffering, loop unrolling, mixed-precision arithmetic) may reduce execution time without compromising numerical fidelity. Coarser integration steps (e.g., $\Delta t = 5$--$20~\text{ms}$) may reveal favorable trade-offs between model fidelity and embedded real-time feasibility.

The framework can be extended to broader probabilistic safety-evaluation tasks. These include closed-loop AEB controller assessment, probabilistic deceleration planning, and rare-event simulation for low-probability high-consequence scenarios.


\section*{Acknowledgment}
AI-assisted tools were used to improve the grammar and wording of portions of this manuscript and to check consistency between the text and the simulation code. All technical content, results, and conclusions are the authors' own.

\bibliographystyle{IEEEtran}
\bibliography{references}

\vspace{2em}
{\footnotesize
\parskip=0pt
\interlinepenalty=500

\noindent\textbf{Akshay Karjol} (Senior Member, IEEE) received the M.S. degree in automotive engineering from Clemson University–International Center for Automotive Research (CU-ICAR), SC, USA. He is currently pursuing the Ph.D. degree in electrical and computer engineering at Oakland University, MI, USA. He is currently with the ADAS department at Ford Motor Company, and has previously held systems engineering and technical leadership roles at ZF Active Safety US Inc. and select automotive OEMs and Tier-1 suppliers, contributing to the development and validation of safety-critical chassis and intelligent vehicle systems, including braking, steering, suspension, advanced driver assistance systems (ADAS), and automated driving systems (ADS), for production vehicle programs. His work spans systems engineering, system architecture, embedded control, and high-performance computing for real-time, safety-critical vehicle applications. His research interests include real-time embedded systems, parallel computing, edge AI, intelligent vehicle systems, and robotics applications.

\vspace{1em}
\noindent\textbf{Shadi Alawneh} (Senior Member, IEEE) received the B.Eng. degree in computer engineering from the Jordan University of Science and Technology, Irbid, Jordan, in 2008, and the M.Eng. and Ph.D. degrees in computer engineering from the Memorial University of Newfoundland, St. John's, NL, Canada, in 2010 and 2014, respectively. Then, he was a Staff Software Developer with the Hardware Acceleration Lab, IBM, Canada, from May 2014 to August 2014. After that, he was a Research Engineer with C-CORE, from 2014 to 2016. He is currently an Associate Professor with the Department of Electrical and Computer Engineering, Oakland University, Rochester, MI, USA. He has authored or coauthored scientific publications, including international peer-reviewed journals and conference papers. His research interests include general-purpose computing on graphics processing units (GPGPU), high-performance computing, embedded system design with GPUs, autonomous driving, numerical simulation and modeling, and software optimization.
\par
}

\end{document}